# An automatic and efficient foreground object extraction scheme

**Subhajit Adhikari[1], Joydeep Kar[2], Jayati Ghosh Dastidar[3]**
[1] St. Xavier's College, Kolkata, India, subhajit15dec@gmail.com
[2] St. Xavier's College, Kolkata, India, jydp.91@gmail.com
[3] St. Xavier's College, Kolkata, India, j.ghoshdastidar@sxccal.edu

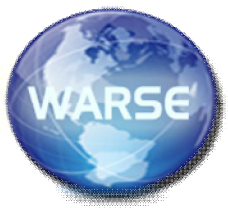

**ABSTRACT**

This paper presents a method to differentiate the foreground objects from the background of a color image. Firstly a color image of any size is input for processing. The algorithm converts it to a grayscale image. Next we apply canny edge detector to find the boundary of the foreground object. We concentrate to find the maximum distance between each boundary pixel column wise and row wise and we fill the region that is bound by the edges. Thus we are able to extract the grayscale values of pixels that are in the bounded region and convert the grayscale image back to original color image containing only the foreground object.

**Key words:** boundary pixel, bounded region, component, edge, foreground object.

## 1. INTRODUCTION

Segmentation is the process that subdivides an image into its constituent parts or objects [1]. Many techniques are proposed to deal with the image segmentation problem such as:

(a) Histogram-Based techniques [1]: this assumes the image to be composed of a number of constant intensity objects in a well-separated background.

(b) Edge-based techniques: where edges are detected and objects are isolated as a result of that.

(c) Region-Based Techniques: where objects are detected according to their homogeneity criteria. Usually, splitting techniques followed by merging ones are involved.

(d) Markov Random Field-Based Techniques: where prior knowledge of the true image together with expensive computations are required.

Image processing systems [1] usually start by an edge detection process followed by a feature extraction technique. Edge detection is implemented to reduce the amount of information in the input image. Feature extraction aims to detect the regions of interest (RoI). The edge detection process reduces the amount of data to be processed but it provides no information about the contents of the image. Thus, further processing is required for the edge-detected-closed-shape objects. Edges define the boundaries [2] between regions in an image, which helps with segmentation and object recognition.

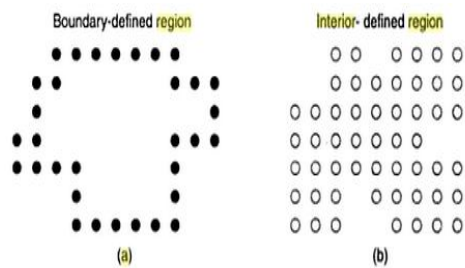

**Figure 1**: Two types of region

Region filling [2] is the process of "coloring in" a definite image area or region. Region may be defined at the pixel or geometric level. At the pixel level, we describe a region either in terms of bounding pixels that outline it or as the totality of pixels that comprise it. The concept of Boundary defined region and Interior defined region [2] is shown above in Figure 1.

In this paper the concept of Edge-based technique is used to detect foreground objects. We propose a simple method that applies the Canny edge detector [3] to detect the boundary of the foreground object. We use the concept of Boundary defined region as a unique value i.e. "1" to fill the region bounded by the edges to form the edge detected image. Then we deal with the actual positions of the boundary pixels. Our aim is to detect the extreme pixels on the boundary and to fill the intermediate region. To do this we just replace the values of the boundary pixels with their positions i.e. firstly with the row number in which the pixel is present fixing the column of that position and vice-versa. Then we compute the intersection of the two approaches and extract the pixels accordingly. Experimentally we have taken RGB color images with simple background and we are able to extract the foreground object in a very short amount of time (sec) using varying image sizes.

## 2. LITERATURE REVIEW

The hexagonal hit-miss transform [1] accepts the input image and produces an edge detected image analyzed by the WST. A novel algorithm called the filling algorithm reads the WST lines and determines the closed shape objects. The mosaic technique is applied to the output of the filling algorithm as to enhance the detection performance.

According to the authors in [2], Canny's method is preferred since it produces single pixel thick, continuous edges.

The authors in paper [3] discussed the most commonly used edge detection techniques of Gradient-based and Laplacian based Edge Detection and under noisy conditions.

In paper [4], all image processing operations generally aim at a better recognition of objects of interest, i.e., at finding suitable local features that can be distinguished from other objects and from the background.

In paper [5], a perfect method for object recognition with full boundary detection by combining affine scale invariant feature transform (ASIFT) and a region merging algorithm is proposed.

A new region growing algorithm is proposed in paper [6], based on the vector angle color similarity measure and the use of the principal component of the covariance matrix as the "characteristic" color of the region, with the goal of a region-based segmentation which is perceptually-based.

In paper [7], the authors propose to extend the watershed segmentation tool to the multi-dimensional case by using the "bit mixing" approach.

In paper [8], two automatic techniques - range data segmentation and camera pose estimation are discussed.

According to the authors in paper [9], the exemplar-based texture synthesis, proposes the approach that employs an exemplar-based texture synthesis technique modulated by a unified scheme for determining the fill order of the target region.

In paper [10], early approaches extended from monochrome edge detection and more recent vector space approaches are addressed.

In paper [11], input image is pre-process to accentuate or remove a band of spatial frequencies and to locate in an image where there is a sudden variation in the grey level of pixels.

**3. METHOD**

Our proposed method is to find the foreground object from a color image. In order to do this we first take a color image of any size & then we convert it to gray scale image. Then we apply canny edge detection. The parameters of canny is set to 0.04 & 0.10 as low and high threshold; and sigma is set to 1.5 [2]. Now we have the edge detected image. Boundary pixels are of value "1" & all other pixels are of value "0". To get the foreground object we have to fill the region bounded by the pixels that are of value "1". We use the simplified idea of Convex hull properties [12], start finding the extreme points scanning boundary pixels row wise and column wise starting from the first occurrence of one boundary pixel. To do this we replace the position of the "1"s with the value of row and find the maximum element without changing the column and vice versa. Thus we are able to find the maximum values corresponding to two directions row wise and column wise and fill the Boundary defined region with the value "1". We extract the pixels accordingly. Thus we have two different representations of foreground objects, one is row wise and another is column wise. We compare and find the intersection of the two representations. We fill the positions and finally extract the pixels from the gray scale image and convert into color image. **Figure 2** above depicts the extraction process.

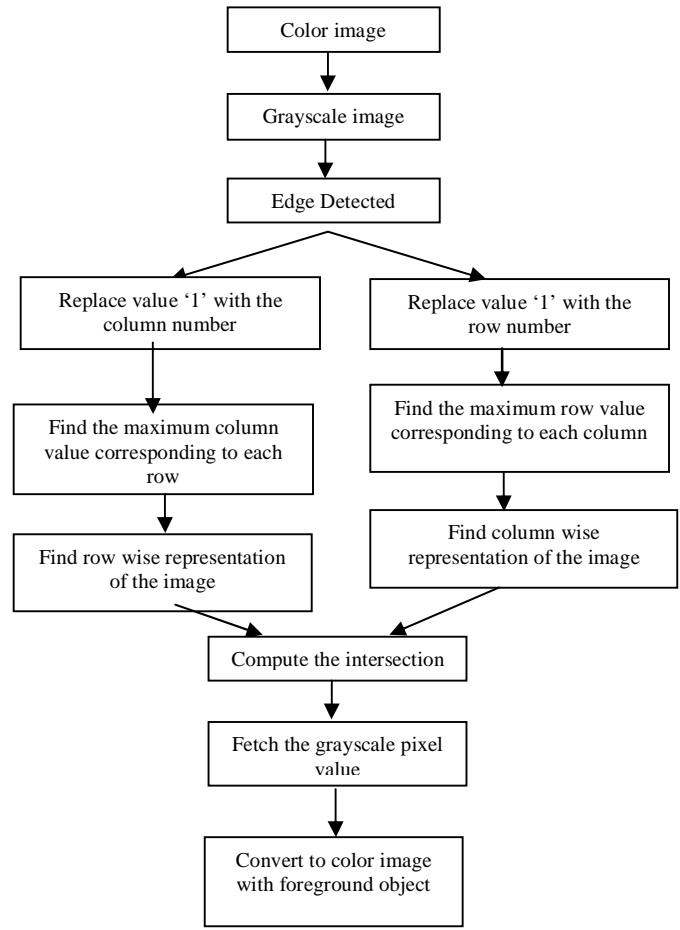

**Figure 2:** Block Diagram for Object Extraction

**4. ALGORITHM**

1. Take a color image of any size and store it to a 2D array.

2. Convert the image into grayscale image and store in a 2D array named, G.

3. Apply canny edge detector with low and high threshold and sigma and save it to matrix BW1 and convert into a 2D matrix, C of type double.

Edge detected image is a binary image that contains "1" and "0" values. Pixels on the boundary are of value "1" and all other pixel values are "0".

Subhajit Adhikari *et al.,* International Journal of Science and Advanced Information Technology, 3 (2), March – April 2014, 40 - 43

4. Replace value '1' with the row number corresponding to each '1' i.e the pixels in the boundary of matrix C and save it a separate matrix D.

5. Find the maximum row value corresponding to each column of the matrix D and save it to an array as k2 (col) .

6. Loop columnwise.

7. Loop rowwise.

8. IF an element of the matrix D is greater than 0
   a. Save the row value of the 1$^{st}$ occurrence of a boundary pixel to a variable k3.
   b. Loop from k3 to k2 (col) i.e. k2 (col) returns maximum row number for the column.
   c. Consider a 2D matrix E and set the value "1" to fill the region inside the boundary pixels.

9. Repeat the steps 4-8.

Obtain matrix D1 that holds the column number in place of each and every boundary pixels that is of value '1'. Then make a loop changing the dimension i.e the row wise and column wise and obtain a separate matrix E1 that holds the region inside the bundary pixels.

10. Compare the two matrices E and E1 to find the intersection portion and extract the portion from the grayscale image and convert into color image to form the foreground object .

**Table 1:** Summary of Execution Time

| Image(.jpg) | Size | Pixels | Time elapsed (in second) |
|---|---|---|---|
| test21 | 324*412 | 133488 | 0.555734146684879 |
| test4 | 376*528 | 198528 | 0.636240589679871 |
| test8 | 438*533 | 233454 | 0.814277172213658 |
| test40 | 600*400 | 240000 | 1.04278616790370 |
| test30 | 422*600 | 253200 | 1.131813784918985 |
| test27 | 424*800 | 339200 | 1.349281825944964 |
| test28 | 600*722 | 433200 | 1.500231680847290 |
| test26 | 800*587 | 469600 | 1.605358465310959 |
| test49 | 1000*768 | 768000 | 1.689927277448329 |
| test 1 | 1050*746 | 783300 | 2.926293942857143 |
| test25 | 1200*797 | 956400 | 4.035249371428572 |
| test2 | 1221*817 | 997557 | 4.603776379525045 |
| test15 | 1366*768 | 1049088 | 4.890780525714286 |
| test17 | 1546*870 | 1345020 | 6.385910857142857 |
| test48 | 1920*1200 | 2304000 | 7.822265498570967 |
| test53 | 2560*1600 | 4096000 | 9.555098898647401 |

## 5. RESULT

We have used MATLAB R2009a for implementing and testing our algorithm. We have used different color images with not very complex background and of different sizes. The table given below (**Table 1**) shows the execution time of our algorithm on some test images. We have run the algorithm on a computer which has a CPU of clock speed 2.8GHz and a RAM of size 2 GB. **Figure 3** shows the effect of our algorithm on some of the test images. The images shown here are not of the original size. We have tried to quantify the deviation in object extraction. We calculated the R-Square and Sum of Squares due to error.

Sum of Squares Due to Error **-** This statistic measures the total deviation of the response values from the fit to the response values. It is also called the summed square of residuals and is usually labeled as SSE (shown below). Here $y_i$ is the observed data value and $\hat{y}_i$ is the predicted value from the fit. $w_i$ is the weighing applied to each data point; usually $w_i = 1$.

$$SSE = \sum_{i=1}^{n} w_i (y_i - \hat{y}_i)^2$$

**Figure 3:** Foreground object extraction

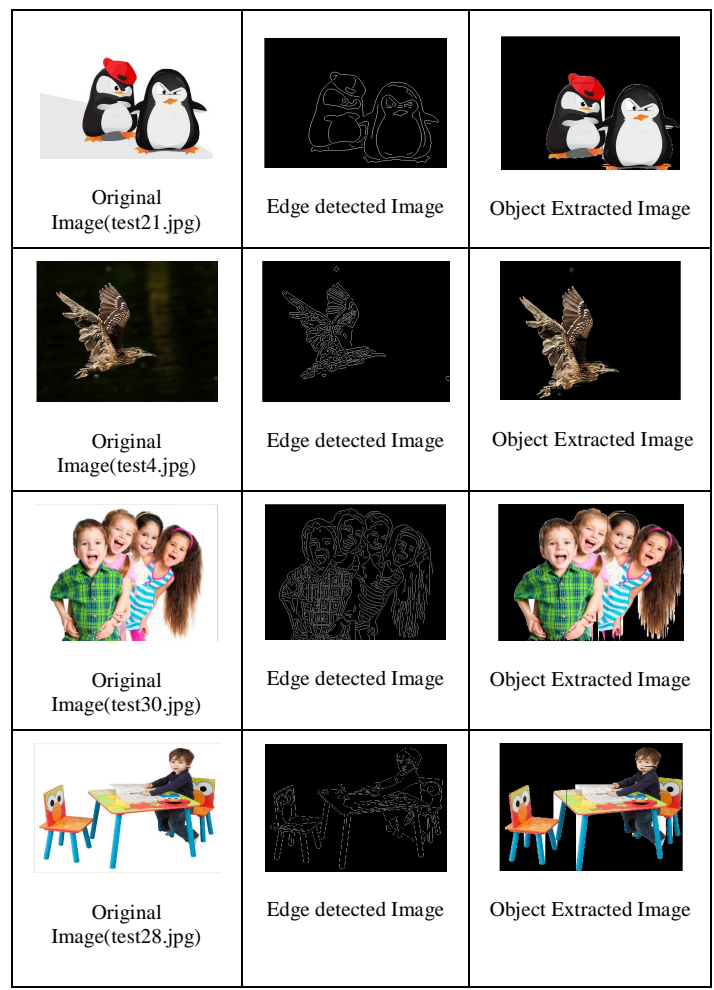

A value closer to 0 indicates that the model has a smaller random error component, and that the fit will be more useful for prediction.

**R-Square** - This statistic measures how successful the fit is in explaining the variation of the data. Put another way, R-square is the square of the correlation between the response values and the predicted response values. It is also called the square of the multiple correlation coefficients and the coefficient of multiple determinations. R-square is defined as the ratio of the sum of squares of the regression (SSR) and the total sum of squares (SST). Here $\hat{y}_i$ is the predicted value from the fit, $\bar{y}$ is the mean of the observed data $y_i$ is the observed data value. $w_i$ is the weighting applied to each data point, usually $w_i=1$. R-square can take on any value between 0 and 1, with a value closer to 1 indicating that a greater proportion of variance is accounted for by the model. For example, an R-square value of 0.8234 means that the fit explains 82.34% of the total variation in the data about the average.

In our test result we get the value of SSE is "0" which means that the model has almost none random error component.

Also the value of the R-square is "1" i.e. the fit explains 100% of the total variation in the data about the average.

residual = data – fit

Residuals are defined as the difference between the observed values of the response variable and the values that are *predicted* by the model. When you fit a model that is appropriate for your data, the residuals approximate independent random errors. We have graphically shown the fit curves in **Figure 4**.

$$SSR = \sum_{i=1}^{} w_i(\hat{y}_i - \bar{y})^2$$

SST is also called the sum of squares about the mean, and is defined as

$$SST = \sum_{i=1}^{n} w_i(y_i - \bar{y})^2$$

where SST = SSR + SSE. Given these definitions, R-square is expressed as

$$R\text{-square} = \frac{SSR}{SST} = 1 - \frac{SSE}{SST}$$

## 6. CONCLUSION

Overall our algorithm takes no human interaction for processing an image. No parameters are to be specified by the user. In future this method can be extended to process color images having more complex background.

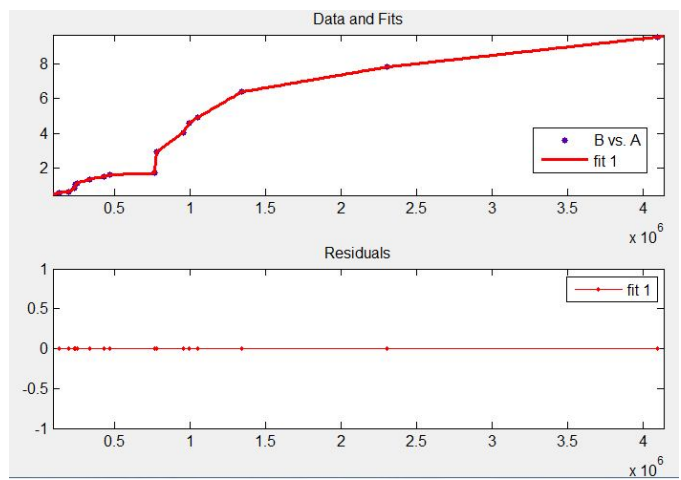

**Figure 4**: Curve Fit